\documentclass[10pt,twocolumn,letterpaper]{article}

\usepackage{multirow}
\usepackage[inline]{enumitem}
\usepackage{cvpr}              

\usepackage[dvipsnames]{xcolor}

\definecolor{cvprblue}{rgb}{0.21,0.49,0.74}
\usepackage[pagebackref,breaklinks,colorlinks,citecolor=cvprblue]{hyperref}

\usepackage[accsupp]{axessibility}
\def\paperID{51} 
\def\confName{CVPR}
\def\confYear{2024}

\title{GestFormer: Multiscale Wavelet Pooling Transformer Network for Dynamic Hand Gesture Recognition }

\author{Mallika Garg, Debashis Ghosh, Pyari Mohan Pradhan \\
Department of Electronics and Communication Engineering, \\ Indian Institute of Technology, Roorkee,  India\\
{\tt\small {mallika@ec.iitr.ac.in, debashis.ghosh@ece.iitr.ac.in, pyarimohan.pradhan@gmail.com}}
}

\begin{document}
\maketitle
\begin{abstract}
Transformer model have achieved state-of-the-art results in many applications like NLP, classification, etc. But their exploration in gesture recognition task is still limited. So, we propose a novel GestFormer architecture for dynamic hand gesture recognition.  The motivation behind this design is to propose a resource efficient transformer model, since transformers are computationally expensive and very complex. So, we propose to use a pooling based token mixer named PoolFormer, since it uses only pooling layer which is a non-parametric layer instead of quadratic attention.  The proposed model also leverages the  space-invariant features of the wavelet transform and also the multiscale features are selected using multi-scale pooling. Further, a gated mechanism helps to focus on fine details of the gesture with the contextual information. This  enhances the performance of the proposed model compared to the traditional transformer with fewer parameters, when evaluated on dynamic hand gesture datasets, NVidia Dynamic Hand Gesture and Briareo datasets. To prove the efficacy of the proposed model, we have experimented on single as well multimodal inputs  such as infrared, normals, depth, optical flow and color images. We have also compared the proposed GestFormer in terms of resource efficiency and number of operations. The source code is available at \url{https://github.com/mallikagarg/GestFormer}.

\end{abstract}    
\section{Introduction}
\label{sec:intro}
Hand gesture recognition is an active and rapidly evolving area of research that involves various applications like sign gesture communication, human-computer interactions, gesture control appliances, autonomous vehicles, virtual reality, gaming etc. This is a  challenging task since it involves variations in the pose, hand shape, position, directions and size of hand.  There are also challenges due to variability of the image background,  color differences, shadows and other lightening illumination which can be handled using depth sensors such as Leap Motion~\cite{potter2013leap} and Microsoft Kinect sensor~\cite{dal2012time}. Gestures can be static or dynamic depending on the movement of hands.  Static hand gestures are those where the hand remains relatively stationary and doesn't involve significant movement while dynamic hand gestures involve movement of the hands or fingers to convey meaning. In this work, we will focus on designing a model that recognizes dynamic hand gestures which are characterized by changes in hand position, orientation, or movement trajectories over time. 

\begin{figure}[tb]				
\centerline{\includegraphics[height=4.5cm, width= 8.7cm]{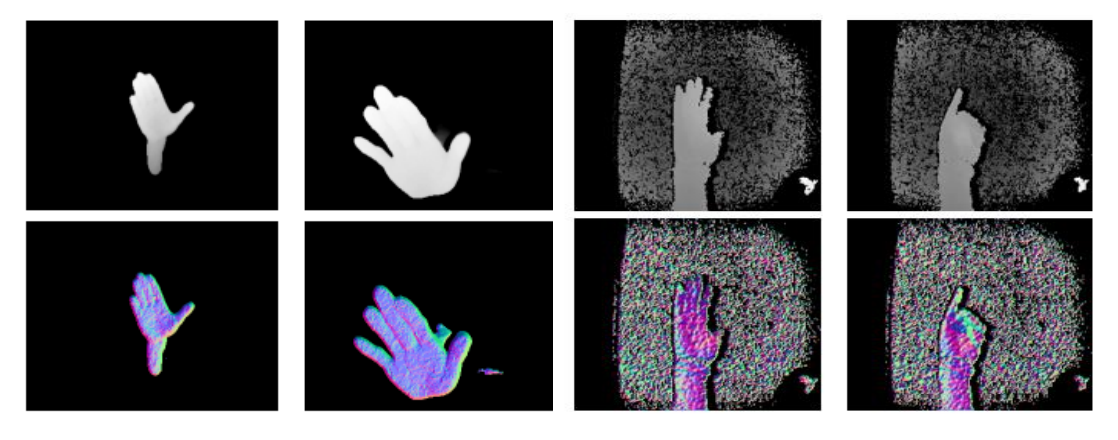}}
\caption{Some samples  of depth (first row) and surface normals (last row) from the NVGesture and Briareo dataset.These samples are taken from~\cite{d2020transformer}.}
\label{fig10}
\end{figure}

With recent advancements in the deep learning algorithms, attention-based models have become popular in focusing on a certain portion of gesture image or video sequence. These attention-based models~\cite{aloysius2021incorporating} have replaced traditional Recurrent Neural Networks (RNNs)~\cite{khodabandelou2020attention}, Long Short-Term Memory (LSTMs)~\cite{avola2018exploiting} and various deep learning methods~\cite{mallika2022two, 9691523} for hand gesture recognition. The recently introduced transformers are one such model that uses attention to focus on a certain portion of image or video sequence. A transformer-based model that classifies dynamic Hand Gesture Recognition is proposed in~\cite{d2020transformer}. This method uses vanilla transformer~\cite{vaswani2017attention}, which comprises of two operation. First, the attention operation is performed which is followed by the  multi-layer perceptron (MLP). The attention operation models the relations between elements while the MLP is employed to model the relation within each individual element. Despite their effectiveness in these domains, their application to visual data, especially in dynamic hand gesture recognition tasks, remains relatively very limited.

So, we explore a transformer-based approach for dynamic gesture recognition. Some samples of the Dynamic gestures from  NVGesture~\cite{molchanov2016online} and Briareo~\cite{manganaro2019hand} dataset are shown in  Fig.~\ref{fig10}. Traditional Transformer~\cite{vaswani2017attention} takes the advantage of quadratic attention which is computationally expensive, $O(n^2)$.  This problem was addresses by Linformer~\cite{wang2020linformer}, which uses linear attention with $O(n)$ complexity in both time and space. With advancements, the attention layer has been completely replaced by layers or modules that has no learnable parameters. PoolFormer~\cite{yu2022metaformer} and  FNet~\cite{lee2021fnet}  proposes an attention-free network which uses average pooling and Fourier transform to mix the token of the input sequence. This helps reduces the complexity of the model to a great level. Inspired from PoolFormer, we also proposed a poolformer based technique that  completely eliminates the attention mechanism and rely on token mixing. Pooling the input can aggregate token from input to learn contextual information and perform comparable to the Vision Transformer with very less complexity.

To further, enhance the performance of the poolformer for dynamic gesture recognition, we propose a novel \textbf{M}ultiscale \textbf{W}avelet \textbf{P}ooling \textbf{A}ttention (MWPA) mechanism  which takes the advantage of wavelet transform~\cite{zhuang2022waveformer} and can be used as an attention approximation mechanism. We also proposed a \textbf{G}ated \textbf{F}eed \textbf{F}orward \textbf{N}etwork (GFFN) to control the flow of the information through the different stages of the proposed \textbf{M}ultiscale \textbf{W}avelet \textbf{P}ooling  \textbf{T}ransformer (MWPT).

Thus, we summarize our key innovations as:
\begin{enumerate}
\item We propose a novel GestFormer, a multiscale wavelet pooling transformer (MWPT) model for dynamic hand gesture recognition.

\item We  propose a novel token mixer called Multiscale Wavelet Pooling Attention (MWPA) which uses multiscale pooling and a wavelet transform to map the input to wavelet space before passing it through the pooling layer. This helps boosts the long-range understanding capabilities of the model.
		
\item  We also propose a Gated Feed forward network which helps to precisely filter the information forwarding to subsequent stages of the transformer block. 

\item Experiments on NVGesture and Briareo dataset are done to prove the efficacy in terms of performance and resource utilisation of the proposed model.

\end{enumerate}

\section{Related Work}
\label{sec:formatting}
In the literature, there are several techniques that rely on traditional methods for hand gesture recognition which often involve manual engineering of features  extraction and the use of classical machine learning algorithms. Earlier hand-crafting features were extracted from raw data, such as images or depth maps of hand gestures to train classical machine learning algorithms such as Support Vector Machines (SVM)~\cite{agarwal2015hand}, Bayesian-classifier~\cite{kumar2018independent},  Hidden Markov Models (HMMs)~\cite{kumar2017coupled}, etc. With these traditional methods for hand gesture recognition, there are issues like robustness, scalability, and adaptability to diverse environments and user conditions that reduces the  performance of the traditional methods. 

\begin{figure*}[tb]				
\centerline{\includegraphics[scale=.55]{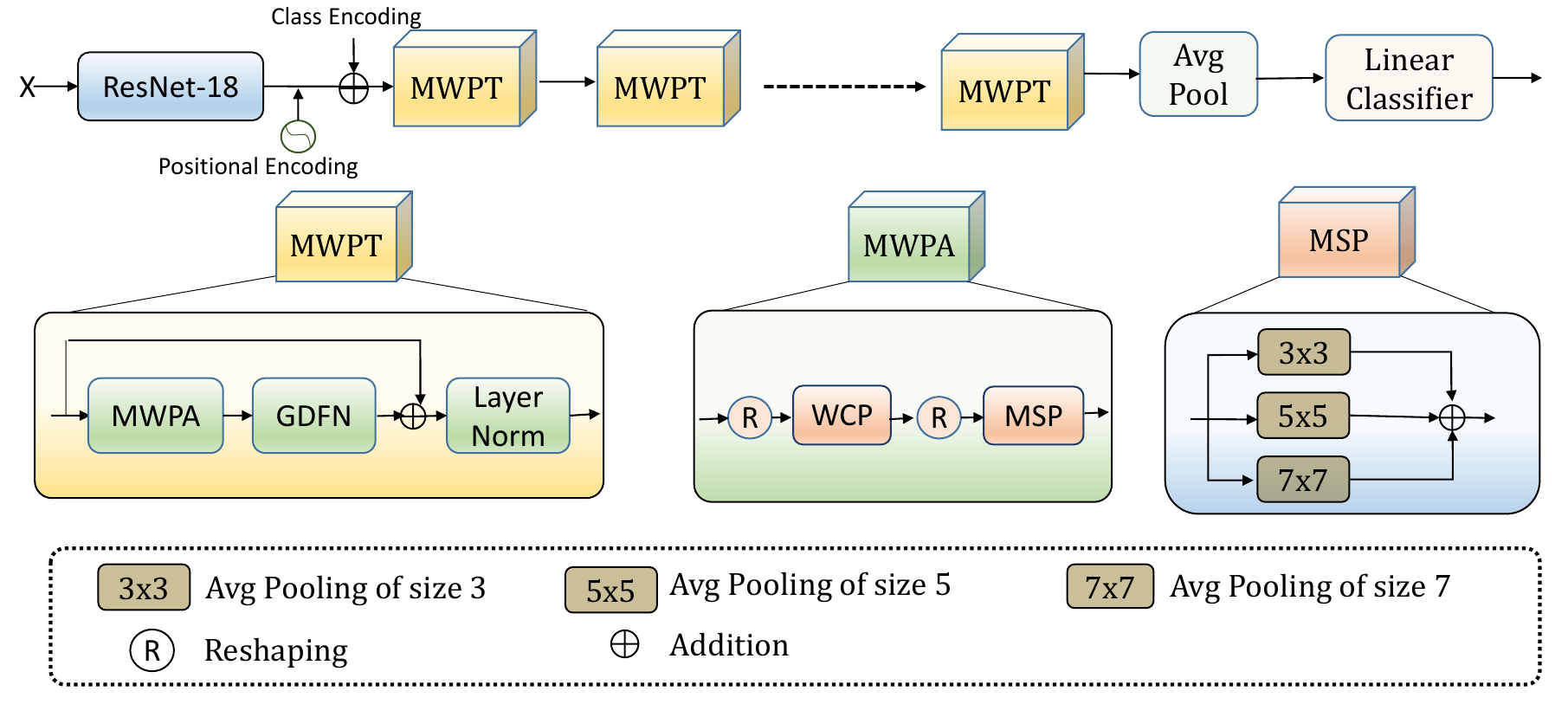}}
\caption{The overall architecture of the proposed GestFormer for dynamic hand gesture recognition. The proposed GestFormer consists of Multiscale Wavelet Pyramid Attention module which comprises of Wavelet Coefficient Processing (WCP) and Multi-scale Pooling architecture (MSP) to perform learning in the wavelet coefficient space with multiscaled pooling to capture the scaled attentive information.  GestFormer also leverage the Gated Dconv FFN (GDFN) to control the forward flow of the information.}
\label{fig1}
\end{figure*}

Later,  with the advent of deep learning, there has been a shift towards more data-driven approaches that automatically learn features from raw data, leading to significant improvements in performance and robustness of the system. With advanced deep learning technologies, Recurrent Neural Networks (RNNs)~\cite{khodabandelou2020attention} and Long Short-Term Memory (LSTMs)~\cite{avola2018exploiting} were developed for handling continuous sign gestures. Nowadays, transformer models are used for gesture recognition, which are designed for sequential data~\cite{de-coster-etal-2020-sign}.

\subsection{Transformer for Vision Tasks}
Transformer-based  networks have shown remarkable success in the field of natural language processing~\cite{vaswani2017attention}, computer vision tasks and modelling sequential data. Since transformers rely on attention mechanism, these models have shown huge progress in  object detection~\cite{zhou2022centerformer}, text generation~\cite{wang2022task}, image classification~\cite{jaderberg2015spatial, chen2022adaptformer, yu2022metaformer}, segmentation~\cite{caron2024location}, recommendation systems~\cite{spurlock2024chatgpt},  super-resolution~\cite{jeevan2024wavemixsr}, dialogue system~\cite{yang2021ubar}, pose estimation~\cite{park2022handoccnet}, text understanding~\cite{atienza2021vision} and many more.  Development of  ViT~\cite{dosovitskiy2020image} marked a significant milestone in the utilization of transformers for vision-based tasks. ViT is a pure transformer-based convolution-free approach which achieves competitive performance compared to CNNs. Later, transformers were used for video based tasks~\cite{neimark2021video}. 

Inspired by ViT, a Video Vision Transformer (ViViT)~\cite{arnab2021vivit} has been introduced that extracts token from video sequence. ViViT presents variants of the models to factorise the spatio-temporal dimensions of the input video: Spatio-temporal attention, Factorised encoder, Factorised dot-product encoder and  Factorised dot-product attention. All these models factorise different components of the transformer model to factorise large spatio-temporal token in the video sequence. DeepViT~\cite{zhou2021deepvit} is another vision transformer which elaborates the issue that attention map goes similar as transformer digs deeper. This signifies that the self attention mechanism fails at deeper layers. So, DeepVit found a solution to this problem by re-generating the attention to get more diverse attention in the deeper layers. 

Although, transformers have marked incredible progress in vision based tasks, they face certain difficulty when these models deal with large sequential data. Since, transformers use quadratic attention, and vision transformers used large sequence length of image tokens, transformers used in visions are computationally expensive and space complexity is also high. Along with this, the vanilla transformer, outputs a feature map of same dimension at each transformer stage.   To tackle these issue, various models that reduce the dimension of the input sequence progressively in the transformer stages are introduced recently. There have been 2 ways to reduce the dimensionality, Convolution based reduction and pooling based reduction. Pyramid Vision Transformer (PvT)~\cite{wang2021pyramid}, Pyramid Pooling Transformer (P2T)~\cite{wu2022p2t}, MsMHA-VTN~\cite{garg2023multiscaled}, MViT~\cite{fan2021multiscale}, Improved MViT~\cite{li2112improved}, PSViT~\cite{chen2021psvit}, POSTER~\cite{zheng2023poster}   are some methods which use pooling to reduce the sequence length and reduce the computation cost of the entire system.  There are some other methods that use pyramid hierarchy but they incorporate convolutional layers instead of pooling e.g.~POTTER~\cite{zheng2023potter},  Convolutional Vision Transformer (CvT)~\cite{wu2021cvt}, Swin Transformer\cite{liu2021swin}, CSwin~\cite{dong2022cswin}, CeiT~\cite{yuan2021incorporating}, Unifying  CNNs~\cite{li2023uniformer}, CoFormer~\cite{deshmukh2024textual}, etc.

\subsection{Token Mixing}
Since, the computation cost of quadratic attention is very high, researchers are now more inclined to replace this attention with some low computational token mixing. PoolFormer~\cite{yu2022metaformer} exploits a general pooling non-parameteric operator to help in basic token mixing. It is the MetaFormer which is actually a generalised mixer for token in computer vision tasks. Another model that mixes the input token by linear transformations (Fourier transform)  is FNet~\cite{lee2021fnet}.  Convolution can also be used to mix tokens as in ConvMixer~\cite{trockman2022patches}. Wavemix~\cite{jeevan2024wavemixsr} uses wavelet transformer and convolution. Similarly, MLP-Mixer~\cite{tolstikhin2021mlp} presents a method that uses MLP for mixing tokens. It separates the channel-mixing and token-mixing task and both tasks use MLP in this architecture.  All these token mixing architectures have comparable performance when compared with the transformer model with less computational requirements.

\subsection{Transformer for Gesture Recognition}
Transformers have nowadays been used in gesture recognition. In~\cite{d2020transformer}, RGBD data is used to predict the class of dynamic gesture using color image , depth maps. It particularly shows that depth maps and the normals which are derived from depth map outperforms other modalities. This method also leverages single and multimodal inputs using basic transformer model. To give the model the order of sequence, positional embedding is employed. An advancement over sinusoidal positional encoding is proposed using  a new positioning scheme based on Gated Recurrent Unit (GRU) into Transformer networks~\cite{aloysius2021incorporating}.

Earlier, multimodal output was taken by fusing the output probability of single modal inputs using decision level fusion, but multimodal fusion at inputs can also be done at the feature fusion stage. One such method~\cite{hampiholi2023convolutional} which  uses convolutional transformer blocks to fuse at the input level is called early fusion. It also performs experiments with mid fusion, late fusion and multi-level fusion.   Spatio-temporal features can also be extracted using transformer models using transformations to canonical maps from both spatial and temporal information~\cite{Cao_2017_ICCV}. Transformer uses columnar structure to map input to same dimensional features. MsMHA-VTN~\cite{garg2023multiscaled} maps the input to multidimensional subspace using pyramid attention networks. This also helps in the reduction of the computational cost of the model. A combined spatiotemporal vision and spatiotemporal channel attention mechanisms can extract  context information from the  input feature  using self attention~\cite{s22062405} on multimodal RGBD data.

\section{Method}

\subsection{Overview}\label{formats}
We propose a transformer-based gesture recognition framework that is designed for dynamic sequence of hand gesture. An overview of the proposed GestFormer is shown in Figure~\ref{fig1}. GestFormer takes a sequence of $m$ frames as an input which can be represented as $X = \{x_{1}, x_{2},..,x_{m}\}$, $X \in \mathbb{R}^{m \times w \times h \times c}$, where $w \times h$ is the size of each frame with $c$ channels. First, the features, $F$ are extracted from each frame using a ResNet-18~\cite{he2016deep} model which outputs a map of $\mathbb{R}^{m \times k}$. These features are then fed to the proposed GestFormer block to learn the wavelet of  multiscale features. Our proposed GestFormer consists of 6 stages of  Multiscale Wavelet Pooling Transformer (MWPT) blocks to get the refined features which finally helps to predict the  probability distribution of $n$ classes using a linear classifier. 

\subsection{Multiscale Wavelet Pooling Transformer (MWPT)} \label{4}
In traditional transformers~\cite{vaswani2017attention}, input is projected into three different vectors, Query, Key, and Value using linear transformation. The attention from these 3 vectors is computed using  scaled-dot product of the Query and Key, normalising it and applying softmax to obtain the weights of the value.  Computation of the attention in this transformer has  quadratic complexity which increase with long sequences. To deal with this issue, we use PoolFormer~\cite{yu2022metaformer} as the core architecture of our proposed MWPT model. PoolFormer replaces the attention mechanism with pooling based token mixing which is a simple non-parametric operation and it has fewer parameters compared to the traditional transformer.

The goal is to develop a model that is computationally less expensive and at the same time, performance of the model is also comparable. PoolFormer achieves competitive results on dynamic gesture recognition when initial experiments were performed. To further enhance the performance, we explore various techniques built on the core PoolFormer structure. The features obtained from the ResNet are first embedded using spatial embedding~\cite{jeevan2024wavemixsr}. We also use positional embedding to make the model know the order of the sequence~\cite{vadisaction}. This encoded input with positional embedding is fed to the proposed MWPT blocks. We propose a novel token mixer called Multiscale Wavelet Pooling Attention (MWPA) which uses multiscale pooling and a wavelet transform before passing the input through the pooling layer. Our MWPA is purely convolution based architecture. After the tokens are mixed in the pool token mixer, we fed the features to the  Gated Depthwise Feed Forward Network (GDFN) block, which helps in selectively passing the fine details in addition with the skip connection to the next stage after layer normalisation. A stack of 6 MWPA stages is used in the proposed MWPT. 

\subsubsection{Multiscale Wavelet Pooling Attention (MWPA)}\label{40}
\begin{figure}[tb]				
\centerline{\includegraphics[scale=.45]{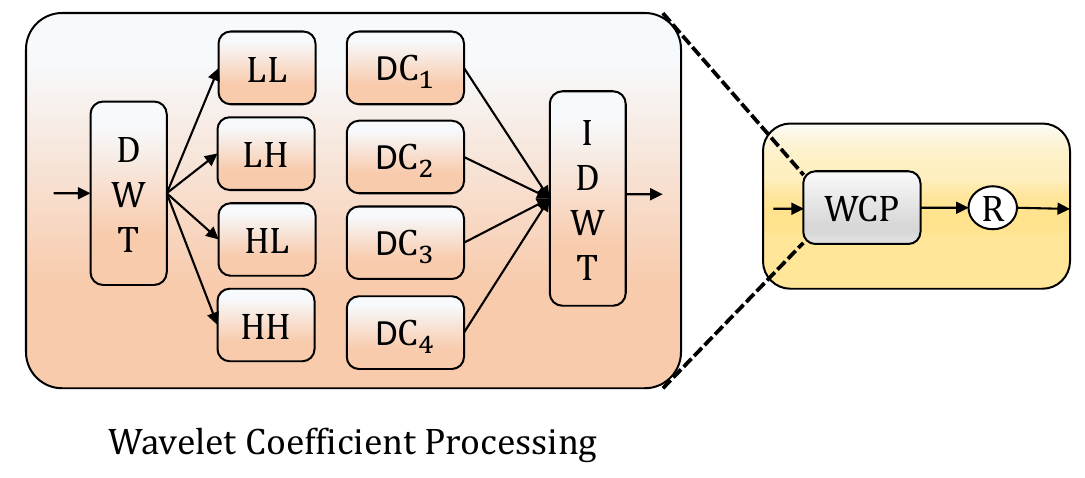}}
\caption{Detailed Wavelet transform Processing (WCP) block. The proposed WCP requires only linear time complexity. It first maps the input into its wavelet coefficients which decomposes the input into multipple sinusoidal waves. The wavelet coefficients is the magnitude of the sinusoidal. After enhancing these magnitudes using Dconv ($DC_x$, x= 1,2,3,4), the coefficients are re-mapped in input space via backward wavelet transform.}
\label{fig2}
\end{figure}

The PoolFormer uses a single input, unlike the vanilla transformer which uses 3 attention vectors. Since the input is fed to the pooling layer directly, it plays an important role in the full transformer block. Pooling helps to select the important features from the input. Further, providing enhanced features as input to the pooling layer can help the model to improve the performance. The enhanced features are calculated by using a wavelet-based forward and backward paradigm~\cite{zhuang2022waveformer}. This facilitates the pooling layer to aggregate the enhanced features in wavelet coefficient space. We follow~\cite{phutke2023blind}, which uses wavelet-based query for image inpainting to reduce the noise forwarding to the attention block. Applying wavelet transform has linear complexity in contrast to transformers which has quadratic complexity. Our model is still less complex.  

\begin{figure}[tb]				
\centerline{\includegraphics[scale=.47]{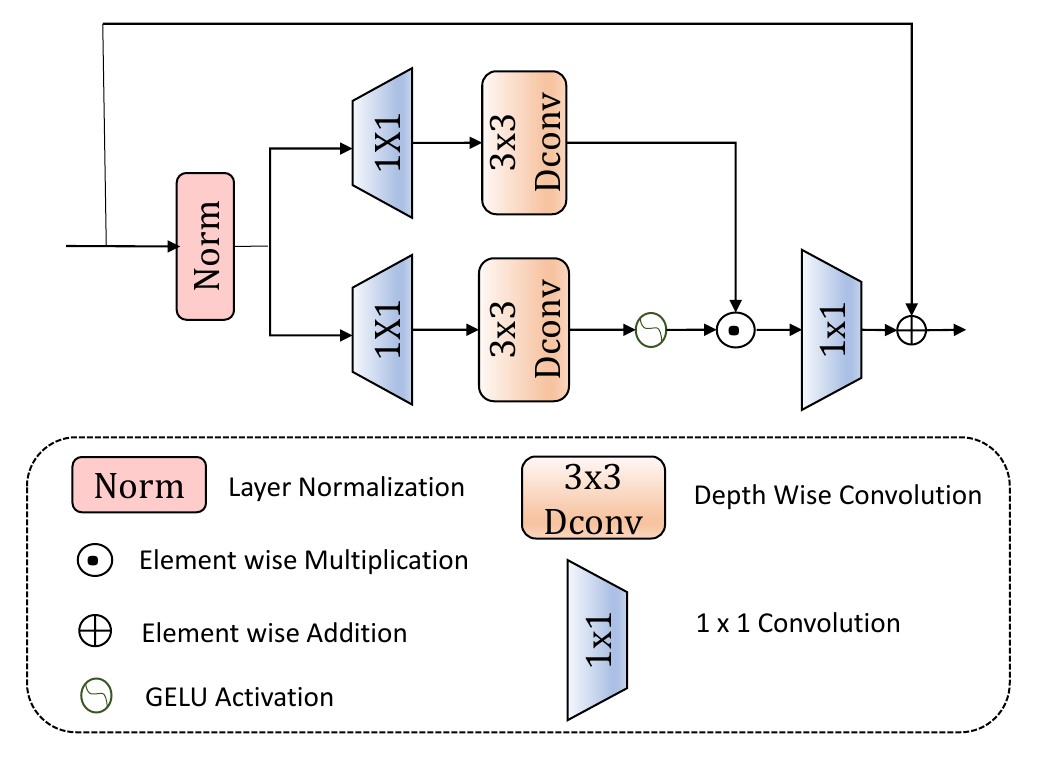}}
\caption{ The detailed Gated Depth-wise Feed Forward Network (GDFN) structure. GDFN facilitates subsequent layers within the network hierarchy to concentrate on more detailed image attributes, thereby resulting in better performance of the complete model.}
\label{fig3}
\end{figure}
We calculate the wavelet coefficient of the input features, $F$ as:

\begin{equation}
\label{eqn:1}	
LL,LH,HL,HH= DWT(F),
\end{equation}

where, the input feature is divided into 4 subspace, which are approximation (LL) and details in 3 orientations as horizontal (LH), vertical (HL) and diagonal (HH). These coefficients are the magnitude of the corresponding sinusoidal wave decomposed after wavelet transform. After the extraction of these coefficients, we separately enhance these features using Depth-wise separable convolution as shown in Fig.~\ref{fig2}.  Further, inverse wavelet transform is calculated from the processed output features, which are given as input to the pooling layer of the PoolFormer. 

In order to extract the important features using pooling, we propose a Multiscale Pooling (MSP) mechanism which helps to aggregate the multiscale information (shown in MSP block in Fig.~\ref{fig1}). A multiscale pooling can accurately capture the hand shape and size variations and recognise the hand with different scales. We propose to apply 3 filters for pooling the input features from the WCP block, ($3 \times 3$, $5\times5$ $7\times 7$). Output from these 3 pooling layers are then averaged to get a strong aggregated enhanced feature. This enhanced feature is the overall output of the proposed MWPA block.

\begin{table*}[tb]
\caption{Results for different modalities on NVGesture~\cite{molchanov2016online}  and Briareo~\cite{manganaro2019hand} dataset. \# is the number of input modalities used. Transformer results are the one reported in~\cite{d2020transformer}.}
\centering
\begin{tabular}{c|ccccc|cc|cc}
\hline
\multirow{3}{*}{\#}&\multicolumn{5}{c|}{Input data}& \multicolumn{4}{c}{Accuracy} \\ \cline{2-10}
&\multirow{3}{*}{Color} &\multirow{3}{*}{ Depth}& \multirow{3}{*}{IR} & \multirow{3}{*}{Normals} & \multirow{3}{*}{Optical flow} & \multicolumn{2}{c|}{NVGesture} & \multicolumn{2}{c}{Briareo}\\  \cline{7-10}
&&&&&&Transformer~\cite{d2020transformer} & {GestFormer} & Transformer~\cite{d2020transformer} & {GestFormer}\\

\hline

\multirow{4}{*}{1} 
&\checkmark&&&&&    76.50\%&75.41\%& 90.60\%&94.44\%  \\
&&\checkmark&&&&    83.00\%&80.21\%& 92.40\%&96.18\%\\
&&&\checkmark&&&    64.70\%&63.54\%& 95.10\%&\textbf{98.13}\%\\
&&&&\checkmark&&    82.40\%&\textbf{81.66}\%& 95.80\%&97.22\%\\ \hline 

\multirow{9}{*}{2}
&\checkmark&\checkmark&&&&  84.60\% &82.57\%&94.10\%&96.78\%\\
&\checkmark&&\checkmark&&&  79.00\%&77.19\%&95.50\%&95.87\%\\
&&\checkmark&\checkmark&&&  81.70\%&79.88\%&95.10\%&96.57\%\\
&\checkmark&&&\checkmark&&  84.60\%&82.75\%&96.50\%&97.44\%\\
&&\checkmark&&\checkmark&&  87.30\%&\textbf{82.78}\%&96.20\%&96.33\%\\
&&&\checkmark&\checkmark&&  83.60\%&82.18\%&97.20\%&\textbf{97.57}\%\\ 
&\checkmark&& &&\checkmark& 72.00\% &72.61\%&-&96.57\%\\
\hline

\multirow{4}{*}{3} 
&\checkmark&\checkmark&\checkmark&&& 85.30\% &\textbf{84.24}\%&95.10\%&96.78\%\\
&\checkmark&\checkmark&&\checkmark&& 86.10\%&83.81\%&95.80\%&\textbf{97.42}\%\\
&\checkmark&&\checkmark&\checkmark&& 85.30\%&83.40\%&96.90\%&96.88\%\\
&&\checkmark&\checkmark&\checkmark&& 87.10\%&83.61\%&97.20\%&96.79\%\\ 
\hline

\multirow{4}{*}{4}
&\checkmark&\checkmark&\checkmark&\checkmark&&  87.60\%&\textbf{85.62}\%&96.20\%&96.33\%\\
&\checkmark&\checkmark&&\checkmark&\checkmark&  -&85.85\%&-&96.79\%\\ 
&\checkmark&&\checkmark&\checkmark&\checkmark&  -&85.31\%&-&\textbf{97.42}\%\\ 
&\checkmark&&\checkmark&\checkmark&\checkmark&  -&84.55\%&-&96.79\%\\ 
&&\checkmark&\checkmark&\checkmark&\checkmark&  -&85.96\%&-&96.79\%\\

\hline
5&\checkmark&\checkmark &\checkmark&\checkmark&\checkmark& -&\textbf{85.85}\%&-&96.88\%\\
\hline

\end{tabular}
\label{tab1}
\end{table*}

\subsubsection{Gated Depthwise Feed Forward Network (GDFN)}\label{5}
To transform the features from MWPA block, we follow~\cite{zamir2022restormer} to apply two modifications in FFN:  gating mechanism and depth-wise convolutions. The architecture of GDFN is shown in Fig.~\ref{fig3}, which helps control the flow of important feature or fine information to the next stage of the transformer blocks. This is formulated by linearly transforming input using depth-wise convolution and performing element-wise product of two parallel features, of which one is Gelu activated represented as. 
\begin{equation}
\label{eqn:2}	
\textbf{{P'}}= W^0_p Gating (\textbf{P}) + \textbf{P},
\end{equation}
\begin{equation}
\label{eqn:3}	
Gating(\textbf{P}) =\phi(W^1_d W^1_p(\textbf{P})) \odot W^2_d W^2_p(\textbf{P}))
\end{equation}
here, $ \odot$ denotes the element-wise multiplication and $\phi $ represents the GELU activation. 

\subsection{Multi-Modal Late Fusion}\label{9}
Multi-modal methods have gained the popularity among research community and have been used in numerous application. RGB-D sensors provides RGB images, depth images, infrared images and it has been used to acquire the NVGestures and Briareo dataset for dynamic hand gesture recognition. Following~\cite{d2020transformer}, we also adapt late fusion technique to predict the multimodal accuracy of the inputs. We have simply averages the output probability score from each input modality trained separately which is given as

\begin{equation}
\label{eqn:6}	
y = \arg\max_j \sum_{i}^n P(\omega_j|x_i),
\end{equation}
where $n$ is the number of modalities over which the results are to be aggregated, and  $P(\omega_j|x_i)$ is the probability distribution of the  $i^{th}$ frames of a given input, which belongs to class $\omega_j$.
\section{Experiments and Discussion}\label{6}
Experiments are performed on single as well as multimodal inputs on  NVGesture and Briareo. We also analyse the number of learnable parameters and the MACs along with the ablation on each component of the model. 

\subsection{Datasets}\label{77}
\textbf{NVGesture:}
NVGesture~\cite{molchanov2016online} is a dynamic hand gesture dataset containing 1582 images in total from 25 different classes. Dataset is divided into two parts, having 1050 samples in training and rest in test dataset. Dataset samples were collected in three different modalities (RGB, IR, and depth) by a group of 20 subjects.

\begin{table}
\caption{Comparison results for single modality on NVGesture dataset~\cite{molchanov2016online}. All results are taken from the respective papers. }
\centering
\begin{tabular}{ccc}
\hline
Input modality&Method& Accuracy \\ 
\hline
\multirow{15}{*}{Color} &Spat. st. CNN~\cite{simonyan2014two} &54.60\%  \\
&iDT-HOG~\cite{wang2016robust}& 59.10\%\\
&Res3ATN~\cite{dhingra2019res3atn}& 62.70\%\\
&C3D~\cite{tran2015learning}&69.30\%\\ 
&R3D-CNN~\cite{molchanov2016online}& 74.10\%\\
&GPM~\cite{fan2021multi}&75.90\% \\
&PreRNN~\cite{yang2018making}&76.50\% \\
&Transformer~\cite{d2020transformer}& 76.50\%\\
&I3D~\cite{wang2016robust}&78.40\%\\
&ResNeXt-101~\cite{kopuklu2019real}& 78.63\%\\
&MTUT~\cite{abavisani2019improving}&81.33\%\\
&NAS1~\cite{yu2021searching}&83.61\% \\
&Human~\cite{molchanov2016online}&88.40\% \\
&MotionRGBD~\cite{zhou2022decoupling}  & 89.57\% \\
&\textbf{GestFormer}& \textbf{75.41\%}\\
\hline

\multirow{12}{*}{Depth}& SNV~\cite{yang2014super}& 70.70\%\\
&C3D~\cite{tran2015learning}&78.80\%\\ 
&R3D-CNN~\cite{molchanov2016online}& 80.30\%\\
&I3D~\cite{wang2016robust}&82.30\%\\
&Transformer~\cite{d2020transformer}&83.00\%\\

&ResNeXt-101~\cite{kopuklu2019real}& 83.82\%\\
&PreRNN~\cite{yang2018making}&84.40\% \\
&MTUT~\cite{abavisani2019improving}&84.85\%\\

&GPM~\cite{fan2021multi}&85.50\% \\
&NAS1~\cite{yu2021searching}&86.10\% \\	
&MotionRGBD~\cite{zhou2022decoupling} & 90.62\% \\
&\textbf{GestFormer}& \textbf{80.21\%}\\

\hline	

\multirow{7}{*}{Optical flow}&iDT-HOF~\cite{vadisaction}& 61.80\% \\
&Temp. st. CNN~\cite{simonyan2014two} & 68.00\%\\
&Transformer~\cite{d2020transformer}& 72.00\%\\
&iDT-MBH~\cite{vadisaction} & 76.80\%\\
&R3D-CNN~\cite{molchanov2016online} & 77.80\%\\
&I3D~\cite{wang2016robust} & 83.40\%\\
&\textbf{GestFormer}&\textbf{72.61\%}\\
\hline

\multirow{2}{*}{Normals}&Transformer~\cite{d2020transformer} &82.40\% \\
&\textbf{GestFormer}& \textbf{81.66\%}\\ \hline

\multirow{3}{*}{Infrared}&R3D-CNN~\cite{molchanov2016online}& 63.50\%\\
& Transformer~\cite{d2020transformer}&  64.70\% \\
&\textbf{GestFormer}&\textbf{63.54\%}\\ \hline

\end{tabular}
\label{tab2}
\end{table}	
  
\begin{table}[tb]
\caption{Comparison results for multi-modalities on NVGestures dataset~\cite{molchanov2016online}. }
\centering
\begin{tabular}{ccc}
\hline
Input modality & Method & Accuracy \\ 
\hline

iDT ~\cite{vadisaction}& color + flow & 73.00\% \\
\hline
R3D-CNN~\cite{molchanov2016online} & color + flow & 79.30\%\\
R3D-CNN~\cite{molchanov2016online} & color + depth + flow & 81.50\%\\
R3D-CNN~\cite{molchanov2016online} & color + depth + ir & 82.00\%\\
R3D-CNN~\cite{molchanov2016online} & depth + flow & 82.40\%\\
R3D-CNN~\cite{molchanov2016online} & all & 83.80\%\\
\hline
MSD-2DCNN~\cite{fan2021multi}&color+depth&84.00\% \\

\hline
8-MFFs-3f1c\cite{kopuklu2018motion}&color + flow& 84.70\%\\
\hline
STSNN~\cite{zhang2020dynamic}&color+flow& 85.13\%\\
\hline
PreRNN~\cite{yang2018making}& color + depth&85.00\% \\
\hline

I3D~\cite{wang2016robust}& color + depth &83.80\%\\
I3D~\cite{wang2016robust}& color + flow &84.40\%\\
I3D~\cite{wang2016robust}& color + depth + flow &85.70\%\\

\hline
GPM~\cite{fan2021multi}& color + depth&86.10\% \\
\hline
MTUT\textsubscript{RGB-D}~\cite{abavisani2019improving}& color + depth& 85.50\%\\
MTUT\textsubscript{RGB-D+flow}~\cite{abavisani2019improving}& color + depth& 86.10\%\\
MTUT\textsubscript{RGB-D+flow}~\cite{abavisani2019improving}& color + depth + flow& 86.90\%\\
\hline

Transformer~\cite{d2020transformer}& depth + normals &87.30\%\\
\multirow{2}{*}{Transformer~\cite{d2020transformer}}& color + depth +ir&\multirow{2}{*}{87.60\%}\\
& + normals & \\
\hline
NAS2~\cite{yu2021searching}& color + depth&86.93\% \\
NAS1+NAS2~\cite{yu2021searching} &color + depth&88.38\% \\
\hline
MotionRGBD~\cite{zhou2022decoupling}&RGB + Depth & 91.70\% \\
\hline
\textbf{GestFormer}&\textbf{depth + normals}&\textbf{82.78\%}\\	
\textbf{GestFormer}&\textbf{depth + color + ir }&\textbf{84.24\%}\\ 
\multirow{2}{*}{\textbf{GestFormer}}&\textbf{depth + color + ir }&\multirow{2}{*}{\textbf{85.62\%}}\\ 
&\textbf{ normal}&\\ 
\multirow{2}{*}{\textbf{GestFormer}}&\textbf{depth + color + ir }&\multirow{2}{*}{\textbf{85.85\%}}\\
& \textbf{normal + op}&\\ 
\hline	
\end{tabular}
\label{tab3}
\end{table}
 
\begin{table}[tb]
\caption{Comparison of the results obtained for different modalities on Briareo dataset~\cite{manganaro2019hand}.}
\centering
\begin{tabular}{ccc}
\hline
Method& Tensor sizes  \\ 
\hline
C3D-HG~\cite{manganaro2019hand}& color& 72.20\%\\
C3D-HG~\cite{manganaro2019hand}& depth& 76.00\%\\
C3D-HG~\cite{manganaro2019hand}& ir& 87.50\%\\

LSTM-HG~\cite{manganaro2019hand}&3D joint features &94.40\%\\
\hline
NUI-CNN~\cite{d2020multimodal}& depth + ir& 92.00\%\\
NUI-CNN~\cite{d2020multimodal}& color + depth + ir& 90.90\%\\

\hline
Transformer~\cite{d2020transformer}& normals& 95.80\%\\
Transformer~\cite{d2020transformer}& depth + normals &96.20\%\\
Transformer~\cite{d2020transformer}&ir + normals &97.20\%\\
\hline	
\textbf{GestFormer} & \textbf{ir}&\textbf{98.13} \%\\
\textbf{GestFormer} & \textbf{ir + normals} &\textbf{97.57}\%\\
\hline
\end{tabular}
\label{tab5}
\end{table}

\textbf{Briareo:}
Briareo dataset~\cite{manganaro2019hand} is collected for  dynamic hand gesture recognition. The dataset samples are collected using a RGB camera, depth sensor, and an infrared stereo camera, under natural lighting conditions. Since images are captured in natural lighting, images are dark and of low contrast. The dataset contains 12 different dynamic gestures which were performed by 40 subjects among them 33 were males and 7 were females. Each gesture is performed 3 times by every subject. Thus a total of 120 ($40 \times 3$) sequences of each gesture is collected of at least 40 frames. Randomly, 32 subjects are placed in the train and the validation set and 8 subjects in the test set.

\subsection{Implementation Details}\label{7}
The proposed GestFormer model was implemented, trained and tested using Torch=1.7.1  with 12 GB Nvidia GeForce GTX 1080 Ti  GPU, CUDA 10.1  with cuDNN 8.1.1.  40 frames  of a gesture are given as input to the model to optimise the loss using  Adam optimizer over categorical cross entropy loss. The model is trained with a batch size of 8 at $1e^{-4}$ learning rate which decays after $50^{th}$ and $75^{th}$ epoch. Following ~\cite{d2020transformer}, we use ResNet-18 model as feature extractor which is  pre-trained on the ImageNet dataset~\cite{deng2009imagenet}. Each modality was separately trained, and probability score for each modality is calculated. Late fusion was used to combine different modalities for integration of diverse sources of information for improved performance.

\subsection{Results and Discussion}
\textbf{NVGesture:} 
We follow~\cite{d2020transformer} to performed experiments with single as well as multi-modality. The result compared with the traditional transformer are compared for single and multimodal combinations for NVGesture in Table~\ref{tab1}. The proposed GestFormer achieves the state-of-the-art results with lesser number of parameters. Lesser parameters are the results of the pooling layers used to replace the attention mechanism. From the table, we can observe that GestFormer obtained best result on normals with an accuracy of 81.66\% and nearly similar result is obtained in depth maps. This is because normals are derived from the depth images.  

Further, the accuracy increases when more than one modality is used as input. The results in multimodal approach are obtained using late fusion. When RGB images are fused with normals or depth maps, an increment in the accuracy is seen. It further increases when normals and depth inputs are fused. Among all the combination of 2 modalities, best performance is obtained when normal and depth is fused which is 82.78\%. From the table, it can be clearly seen  that adding a modality shows an increment in the accuracy. 3 modality reaches an accuracy of 84.24\% with RGB, depth and IR fusion. Accuracy further improves to 85.62\% with 4 modal input and the best accuracy is obtained with all the 5 modalities which is 85.85\% on the proposed GestFormer. However, it is still less compared to the traditional transformer~\cite{d2020transformer} on single as well as multimodal inputs.

We also compare the performance of the proposed GestFormer with other methods on single modality in Table~\ref{tab2}, and on multimodal inputs in Table~\ref{tab3} and observe that GestFormer achieves  state-of-the-art results. We can also observe from the  Table~\ref{tab2} that our model is able to outperform Transformer model~\cite{d2020transformer} when optical flow input is given to the model.

\textbf{Briareo:} 
Similar to NVGesture, we performed experiments on Briareo dataset with single and multimodal inputs as shown in table~\ref{tab1}. A comparison is also shown with the basic transformer architecture~\cite{d2020transformer}. From the comparison, we can conclude that  GestFormer performs better on Briareo dataset compared to \cite{d2020transformer} with approx. 2-4\% rise in accuracy with each modality. It can  also be observed that our modal has better results on all the modalities and also on all the experiments individually, except the 2 experiments with 3 modalities. Best performance is observed when infra-red input is used, obtaining an 98.13\% accuracy. Combining modalities did not lead to a notable improvement in GestFormer's performance.

Additionally, we also compared the results obtained by  the proposed model with other methods  in Table~\ref{tab5}. It is evident that GestFormer achieves superior performance with an accuracy of 98.13\%. Finally, we can also conclude from the results that  GestFormer is able to achieve better results on single modalities, leading to a conclusion that even without using multimodal inputs for our methods, we are able to achieve better results than other state-of-the-art methods.

\subsection{Ablation Study}
We perform the ablation study on NVGesture depth modality. The proposed GestFormer has 8 baselines (BL1, BL2, BL3, BL4, BL5, BL6, BL7 and BL8) as shown in Table~\ref{tab25}. Baseline BL1 is the transformer model with pooling layer similar to PoolFormer (A). Baseline BL2   explores the  pooling transformer with the multi-scaling pooling network (B)  where 3 types of filters are used for each scale. Baseline BL3 uses encoding of input using spatial embedding (C) with A as discussed in Section~\ref{4}. Baseline BL4 and BL5 is the Wavelet transform (WCP) (D) and Gated Dconv FFN (GDFN) (E)  used with A. 

An initial experiment that shows the performance of PoolFormer is 76.04\% which increases to 76.67\% by using  multi-scale pooling network. Further, addition of different modules to the poolformer aims to enhance the performance of the proposed model. From the table, we can conclude that addition of each baseline on BL1 has enhanced the performance of the model, giving a clear motivation of designing the proposed GestFormer model. 

We have also compared the number of learnable parameters  and the number of  MAC of our model with other models and the traditional transformer model  in Table~\ref{tab24}. The numbers of parameters and MACs are comparatively less  for GestFormer from other methods.

\begin{table}[tb]
\caption{Ablation study on the proposed GestFormer model.}
\centering
\begin{tabular}{c|c|c }
\hline
Baseline & Module & Accuracy\\  \hline
BL1&PoolFormer (A) & 76.04\\
BL2&A + MSP(B) & 76.67\\
BL3&A + embedding(C) & 77.29\\
BL4&A + WCP(D) & 78.95\\
BL5&A + GDFN(E) & 79.12\\
BL6&A + C + D & 79.58\\
BL7&A + B + C + D & 79.97\\

BL8&A + B + C + D + E & \textbf{80.21}\\
\hline
\end{tabular}
\label{tab25}
\end{table}

\begin{table}[tb]
\caption{Comparison in terms of the number of parameters (M)) and MACs. The numbers
of MACs are counted by fvcore library. }
\centering
\begin{tabular}{c|cc}
\hline
Methods&Params (M)& MACs (G)\\  \hline
R3D-CNN~\cite{molchanov2016online} & 38.00&-\\
C3D-HG~\cite{manganaro2019hand} & 26.70&-\\
Transformer\cite{d2020transformer} & 24.30&62.92\\
\textbf{GestFormer}&\textbf{24.08}&\textbf{60.40}\\
\hline
\end{tabular}
\label{tab24}
\end{table}	

\section{Conclusion}
We proposed a novel GestFormer model for dynamic hand gesture recognition build on PoolFormer which is a computationally efficient model since it uses non parametric layer. We further enhance the performance by extracting wavelet coefficients and enhancing the features in wavelet space. We also leverage the multiscale contextual information by using multiscale pooling and a gated network to process the  refined features. This helps the model to learn significant features with fewer parameters compared to the traditional transformer.  Evaluating the proposed GestFormer on NVGesture and Briareo datasets shows our model achieves state-of-the-art results. For Briareo dataset,  we can conclude that our GestFormer model is so efficient that it performs better with single  input compared to other single and multimodal methods as well.

{
    \small
    \bibliographystyle{ieeenat_fullname}
    \bibliography{main}

\begin{thebibliography}{71}
\providecommand{\natexlab}[1]{#1}
\providecommand{\url}[1]{\texttt{#1}}
\expandafter\ifx\csname urlstyle\endcsname\relax
  \providecommand{\doi}[1]{doi: #1}\else
  \providecommand{\doi}{doi: \begingroup \urlstyle{rm}\Url}\fi

\bibitem[Abavisani et~al.(2019)Abavisani, Joze, and Patel]{abavisani2019improving}
Mahdi Abavisani, Hamid Reza~Vaezi Joze, and Vishal~M Patel.
\newblock Improving the performance of unimodal dynamic hand-gesture recognition with multimodal training.
\newblock In \emph{Proceedings of the IEEE/CVF conference on computer vision and pattern recognition}, pages 1165--1174, 2019.

\bibitem[Agarwal et~al.(2015)Agarwal, Raman, and Mittal]{agarwal2015hand}
Rajat Agarwal, Balasubramanian Raman, and Ankush Mittal.
\newblock Hand gesture recognition using discrete wavelet transform and support vector machine.
\newblock In \emph{2015 2nd International Conference on Signal Processing and Integrated Networks (SPIN)}, pages 489--493. IEEE, 2015.

\bibitem[Aloysius et~al.(2021)Aloysius, Geetha, and Nedungadi]{aloysius2021incorporating}
Neena Aloysius, M Geetha, and Prema Nedungadi.
\newblock Incorporating relative position information in transformer-based sign language recognition and translation.
\newblock \emph{IEEE Access}, 9:\penalty0 145929--145942, 2021.

\bibitem[Arnab et~al.(2021)Arnab, Dehghani, Heigold, Sun, Lu{\v{c}}i{\'c}, and Schmid]{arnab2021vivit}
Anurag Arnab, Mostafa Dehghani, Georg Heigold, Chen Sun, Mario Lu{\v{c}}i{\'c}, and Cordelia Schmid.
\newblock Vivit: A video vision transformer.
\newblock In \emph{Proceedings of the IEEE/CVF international conference on computer vision}, pages 6836--6846, 2021.

\bibitem[Atienza(2021)]{atienza2021vision}
Rowel Atienza.
\newblock Vision transformer for fast and efficient scene text recognition.
\newblock In \emph{International Conference on Document Analysis and Recognition}, pages 319--334. Springer, 2021.

\bibitem[Avola et~al.(2018)Avola, Bernardi, Cinque, Foresti, and Massaroni]{avola2018exploiting}
Danilo Avola, Marco Bernardi, Luigi Cinque, Gian~Luca Foresti, and Cristiano Massaroni.
\newblock Exploiting recurrent neural networks and leap motion controller for the recognition of sign language and semaphoric hand gestures.
\newblock \emph{IEEE Transactions on Multimedia}, 21\penalty0 (1):\penalty0 234--245, 2018.

\bibitem[Cao et~al.(2017)Cao, Zhang, Wu, Lu, and Cheng]{Cao_2017_ICCV}
Congqi Cao, Yifan Zhang, Yi Wu, Hanqing Lu, and Jian Cheng.
\newblock Egocentric gesture recognition using recurrent 3d convolutional neural networks with spatiotemporal transformer modules.
\newblock In \emph{Proceedings of the IEEE International Conference on Computer Vision (ICCV)}, 2017.

\bibitem[Caron et~al.(2024)Caron, Houlsby, and Schmid]{caron2024location}
Mathilde Caron, Neil Houlsby, and Cordelia Schmid.
\newblock Location-aware self-supervised transformers for semantic segmentation.
\newblock In \emph{Proceedings of the IEEE/CVF Winter Conference on Applications of Computer Vision}, pages 117--127, 2024.

\bibitem[Chen et~al.(2021)Chen, Li, Li, Li, Bai, Lin, Sun, Yan, and Ouyang]{chen2021psvit}
Boyu Chen, Peixia Li, Baopu Li, Chuming Li, Lei Bai, Chen Lin, Ming Sun, Junjie Yan, and Wanli Ouyang.
\newblock Psvit: Better vision transformer via token pooling and attention sharing.
\newblock \emph{arXiv preprint arXiv:2108.03428}, 2021.

\bibitem[Chen et~al.(2022{\natexlab{a}})Chen, Li, Fang, Xin, Lu, and Miao]{s22062405}
Huizhou Chen, Yunan Li, Huijuan Fang, Wentian Xin, Zixiang Lu, and Qiguang Miao.
\newblock Multi-scale attention 3d convolutional network for multimodal gesture recognition.
\newblock \emph{Sensors}, 22, 2022{\natexlab{a}}.

\bibitem[Chen et~al.(2022{\natexlab{b}})Chen, Ge, Tong, Wang, Song, Wang, and Luo]{chen2022adaptformer}
Shoufa Chen, Chongjian Ge, Zhan Tong, Jiangliu Wang, Yibing Song, Jue Wang, and Ping Luo.
\newblock Adaptformer: Adapting vision transformers for scalable visual recognition.
\newblock \emph{Advances in Neural Information Processing Systems}, 35:\penalty0 16664--16678, 2022{\natexlab{b}}.

\bibitem[Dal~Mutto et~al.(2012)Dal~Mutto, Zanuttigh, and Cortelazzo]{dal2012time}
Carlo Dal~Mutto, Pietro Zanuttigh, and Guido~M Cortelazzo.
\newblock \emph{Time-of-flight cameras and Microsoft KinectTM}.
\newblock Springer Science \& Business Media, 2012.

\bibitem[De~Coster et~al.(2020)De~Coster, Van~Herreweghe, and Dambre]{de-coster-etal-2020-sign}
Mathieu De~Coster, Mieke Van~Herreweghe, and Joni Dambre.
\newblock Sign language recognition with transformer networks.
\newblock In \emph{Proceedings of the Twelfth Language Resources and Evaluation Conference}, 2020.

\bibitem[Deng et~al.(2009)Deng, Dong, Socher, Li, Li, and Fei-Fei]{deng2009imagenet}
Jia Deng, Wei Dong, Richard Socher, Li-Jia Li, Kai Li, and Li Fei-Fei.
\newblock Imagenet: A large-scale hierarchical image database.
\newblock In \emph{2009 IEEE conference on computer vision and pattern recognition}, pages 248--255. Ieee, 2009.

\bibitem[Deshmukh et~al.(2024)Deshmukh, Susladkar, Makwana, Mittal, et~al.]{deshmukh2024textual}
Gayatri Deshmukh, Onkar Susladkar, Dhruv Makwana, Sparsh Mittal, et~al.
\newblock Textual alchemy: Coformer for scene text understanding.
\newblock In \emph{Proceedings of the IEEE/CVF Winter Conference on Applications of Computer Vision}, pages 2931--2941, 2024.

\bibitem[Dhingra and Kunz(2019)]{dhingra2019res3atn}
Naina Dhingra and Andreas Kunz.
\newblock Res3atn-deep 3d residual attention network for hand gesture recognition in videos.
\newblock In \emph{2019 international conference on 3D vision (3DV)}, pages 491--501. IEEE, 2019.

\bibitem[Dong et~al.(2022)Dong, Bao, Chen, Zhang, Yu, Yuan, Chen, and Guo]{dong2022cswin}
Xiaoyi Dong, Jianmin Bao, Dongdong Chen, Weiming Zhang, Nenghai Yu, Lu Yuan, Dong Chen, and Baining Guo.
\newblock Cswin transformer: A general vision transformer backbone with cross-shaped windows.
\newblock In \emph{Proceedings of the IEEE/CVF Conference on Computer Vision and Pattern Recognition}, pages 12124--12134, 2022.

\bibitem[Dosovitskiy et~al.(2020)Dosovitskiy, Beyer, Kolesnikov, Weissenborn, Zhai, Unterthiner, Dehghani, Minderer, Heigold, Gelly, et~al.]{dosovitskiy2020image}
Alexey Dosovitskiy, Lucas Beyer, Alexander Kolesnikov, Dirk Weissenborn, Xiaohua Zhai, Thomas Unterthiner, Mostafa Dehghani, Matthias Minderer, Georg Heigold, Sylvain Gelly, et~al.
\newblock An image is worth 16x16 words: Transformers for image recognition at scale.
\newblock \emph{arXiv preprint arXiv:2010.11929}, 2020.

\bibitem[D’Eusanio et~al.(2020{\natexlab{a}})D’Eusanio, Simoni, Pini, Borghi, Vezzani, and Cucchiara]{d2020multimodal}
Andrea D’Eusanio, Alessandro Simoni, Stefano Pini, Guido Borghi, Roberto Vezzani, and Rita Cucchiara.
\newblock Multimodal hand gesture classification for the human--car interaction.
\newblock In \emph{Informatics}, page~31, 2020{\natexlab{a}}.

\bibitem[D’Eusanio et~al.(2020{\natexlab{b}})D’Eusanio, Simoni, Pini, Borghi, Vezzani, and Cucchiara]{d2020transformer}
Andrea D’Eusanio, Alessandro Simoni, Stefano Pini, Guido Borghi, Roberto Vezzani, and Rita Cucchiara.
\newblock A transformer-based network for dynamic hand gesture recognition.
\newblock In \emph{International Conference on 3D Vision (3DV)}, pages 623--632. IEEE, 2020{\natexlab{b}}.

\bibitem[Fan et~al.(2021{\natexlab{a}})Fan, Lu, Xu, and Cao]{fan2021multi}
Dinghao Fan, Hengjie Lu, Shugong Xu, and Shan Cao.
\newblock Multi-task and multi-modal learning for rgb dynamic gesture recognition.
\newblock \emph{IEEE Sensors Journal}, 21\penalty0 (23):\penalty0 27026--27036, 2021{\natexlab{a}}.

\bibitem[Fan et~al.(2021{\natexlab{b}})Fan, Xiong, Mangalam, Li, Yan, Malik, and Feichtenhofer]{fan2021multiscale}
Haoqi Fan, Bo Xiong, Karttikeya Mangalam, Yanghao Li, Zhicheng Yan, Jitendra Malik, and Christoph Feichtenhofer.
\newblock Multiscale vision transformers.
\newblock In \emph{Proceedings of the IEEE/CVF international conference on computer vision}, pages 6824--6835, 2021{\natexlab{b}}.

\bibitem[Garg et~al.(2021)Garg, Pradhan, and Ghosh]{9691523}
Mallika Garg, Pyari~Mohan Pradhan, and Debashis Ghosh.
\newblock Multiview hand gesture recognition using deep learning.
\newblock In \emph{2021 IEEE 18th India Council International Conference (INDICON)}, 2021.

\bibitem[Garg et~al.(2023)Garg, Ghosh, and Pradhan]{garg2023multiscaled}
Mallika Garg, Debashis Ghosh, and Pyari~Mohan Pradhan.
\newblock Multiscaled multi-head attention-based video transformer network for hand gesture recognition.
\newblock \emph{IEEE Signal Processing Letters}, 30:\penalty0 80--84, 2023.

\bibitem[Hampiholi et~al.(2023)Hampiholi, Jarvers, Mader, and Neumann]{hampiholi2023convolutional}
Basavaraj Hampiholi, Christian Jarvers, Wolfgang Mader, and Heiko Neumann.
\newblock Convolutional transformer fusion blocks for multi-modal gesture recognition.
\newblock \emph{IEEE Access}, 11:\penalty0 34094--34103, 2023.

\bibitem[He et~al.(2016)He, Zhang, Ren, and Sun]{he2016deep}
Kaiming He, Xiangyu Zhang, Shaoqing Ren, and Jian Sun.
\newblock Deep residual learning for image recognition.
\newblock In \emph{Proceedings of the IEEE conference on computer vision and pattern recognition}, pages 770--778, 2016.

\bibitem[Jaderberg et~al.(2015)Jaderberg, Simonyan, Zisserman, et~al.]{jaderberg2015spatial}
Max Jaderberg, Karen Simonyan, Andrew Zisserman, et~al.
\newblock Spatial transformer networks.
\newblock \emph{Advances in neural information processing systems}, 28, 2015.

\bibitem[Jeevan et~al.(2024)Jeevan, Srinidhi, Prathiba, and Sethi]{jeevan2024wavemixsr}
Pranav Jeevan, Akella Srinidhi, Pasunuri Prathiba, and Amit Sethi.
\newblock Wavemixsr: Resource-efficient neural network for image super-resolution.
\newblock In \emph{Proceedings of the IEEE/CVF Winter Conference on Applications of Computer Vision}, pages 5884--5892, 2024.

\bibitem[Khodabandelou et~al.(2020)Khodabandelou, Jung, Amirat, and Mohammed]{khodabandelou2020attention}
Ghazaleh Khodabandelou, Pyeong-Gook Jung, Yacine Amirat, and Samer Mohammed.
\newblock Attention-based gated recurrent unit for gesture recognition.
\newblock \emph{IEEE Transactions on Automation Science and Engineering}, 18\penalty0 (2):\penalty0 495--507, 2020.

\bibitem[Kopuklu et~al.(2018)Kopuklu, Kose, and Rigoll]{kopuklu2018motion}
Okan Kopuklu, Neslihan Kose, and Gerhard Rigoll.
\newblock Motion fused frames: Data level fusion strategy for hand gesture recognition.
\newblock In \emph{Proceedings of the IEEE conference on computer vision and pattern recognition workshops}, pages 2103--2111, 2018.

\bibitem[K{\"o}p{\"u}kl{\"u} et~al.(2019)K{\"o}p{\"u}kl{\"u}, Gunduz, Kose, and Rigoll]{kopuklu2019real}
Okan K{\"o}p{\"u}kl{\"u}, Ahmet Gunduz, Neslihan Kose, and Gerhard Rigoll.
\newblock Real-time hand gesture detection and classification using convolutional neural networks.
\newblock In \emph{2019 14th IEEE international conference on automatic face \& gesture recognition (FG 2019)}, pages 1--8. IEEE, 2019.

\bibitem[Kumar et~al.(2017)Kumar, Gauba, Roy, and Dogra]{kumar2017coupled}
Pradeep Kumar, Himaanshu Gauba, Partha~Pratim Roy, and Debi~Prosad Dogra.
\newblock Coupled hmm-based multi-sensor data fusion for sign language recognition.
\newblock \emph{Pattern Recognition Letters}, 86:\penalty0 1--8, 2017.

\bibitem[Kumar et~al.(2018)Kumar, Roy, and Dogra]{kumar2018independent}
Pradeep Kumar, Partha~Pratim Roy, and Debi~Prosad Dogra.
\newblock Independent bayesian classifier combination based sign language recognition using facial expression.
\newblock \emph{Information Sciences}, 428:\penalty0 30--48, 2018.

\bibitem[Lee-Thorp et~al.(2021)Lee-Thorp, Ainslie, Eckstein, and Ontanon]{lee2021fnet}
James Lee-Thorp, Joshua Ainslie, Ilya Eckstein, and Santiago Ontanon.
\newblock Fnet: Mixing tokens with fourier transforms.
\newblock \emph{arXiv preprint arXiv:2105.03824}, 2021.

\bibitem[Li et~al.(2023)Li, Wang, Zhang, Gao, Song, Liu, Li, and Qiao]{li2023uniformer}
Kunchang Li, Yali Wang, Junhao Zhang, Peng Gao, Guanglu Song, Yu Liu, Hongsheng Li, and Yu Qiao.
\newblock Uniformer: Unifying convolution and self-attention for visual recognition.
\newblock \emph{IEEE Transactions on Pattern Analysis and Machine Intelligence}, 2023.

\bibitem[Li et~al.(2021)Li, Wu, Fan, Mangalam, Xiong, Malik, and Feichtenhofer]{li2112improved}
Y Li, CY Wu, H Fan, K Mangalam, B Xiong, J Malik, and C Feichtenhofer.
\newblock Improved multiscale vision transformers for classification and detection.
\newblock \emph{arXiv preprint arXiv:2112.01526}, 2021.

\bibitem[Liu et~al.(2021)Liu, Lin, Cao, Hu, Wei, Zhang, Lin, and Guo]{liu2021swin}
Ze Liu, Yutong Lin, Yue Cao, Han Hu, Yixuan Wei, Zheng Zhang, Stephen Lin, and Baining Guo.
\newblock Swin transformer: Hierarchical vision transformer using shifted windows.
\newblock In \emph{Proceedings of the IEEE/CVF international conference on computer vision}, pages 10012--10022, 2021.

\bibitem[Mallika et~al.(2022)Mallika, Ghosh, and Pradhan]{mallika2022two}
Garg Mallika, Debashis Ghosh, and Pyari~Mohan Pradhan.
\newblock A two-stage convolutional neural network for hand gesture recognition.
\newblock In \emph{Proceedings of the 6th International Conference on Advance Computing and Intelligent Engineering: ICACIE 2021}, 2022.

\bibitem[Manganaro et~al.(2019)Manganaro, Pini, Borghi, Vezzani, and Cucchiara]{manganaro2019hand}
Fabio Manganaro, Stefano Pini, Guido Borghi, Roberto Vezzani, and Rita Cucchiara.
\newblock Hand gestures for the human-car interaction: The briareo dataset.
\newblock In \emph{Image Analysis and Processing--ICIAP 2019: 20th International Conference, Trento, Italy, September 9--13, 2019, Proceedings, Part II 20}, pages 560--571. Springer, 2019.

\bibitem[Molchanov et~al.(2016)Molchanov, Yang, Gupta, Kim, Tyree, and Kautz]{molchanov2016online}
Pavlo Molchanov, Xiaodong Yang, Shalini Gupta, Kihwan Kim, Stephen Tyree, and Jan Kautz.
\newblock Online detection and classification of dynamic hand gestures with recurrent 3d convolutional neural network.
\newblock In \emph{Proceedings of the IEEE conference on computer vision and pattern recognition}, pages 4207--4215, 2016.

\bibitem[Neimark et~al.(2021)Neimark, Bar, Zohar, and Asselmann]{neimark2021video}
Daniel Neimark, Omri Bar, Maya Zohar, and Dotan Asselmann.
\newblock Video transformer network.
\newblock In \emph{Proceedings of the IEEE/CVF international conference on computer vision}, pages 3163--3172, 2021.

\bibitem[Park et~al.(2022)Park, Oh, Moon, Choi, and Lee]{park2022handoccnet}
JoonKyu Park, Yeonguk Oh, Gyeongsik Moon, Hongsuk Choi, and Kyoung~Mu Lee.
\newblock Handoccnet: Occlusion-robust 3d hand mesh estimation network.
\newblock In \emph{Proceedings of the IEEE/CVF Conference on Computer Vision and Pattern Recognition}, pages 1496--1505, 2022.

\bibitem[Phutke et~al.(2023)Phutke, Kulkarni, Vipparthi, and Murala]{phutke2023blind}
Shruti~S Phutke, Ashutosh Kulkarni, Santosh~Kumar Vipparthi, and Subrahmanyam Murala.
\newblock Blind image inpainting via omni-dimensional gated attention and wavelet queries.
\newblock In \emph{Proceedings of the IEEE/CVF Conference on Computer Vision and Pattern Recognition}, pages 1251--1260, 2023.

\bibitem[Potter et~al.(2013)Potter, Araullo, and Carter]{potter2013leap}
Leigh~Ellen Potter, Jake Araullo, and Lewis Carter.
\newblock The leap motion controller: a view on sign language.
\newblock In \emph{Proceedings of the 25th Australian computer-human interaction conference: augmentation, application, innovation, collaboration}, pages 175--178, 2013.

\bibitem[Simonyan and Zisserman(2014)]{simonyan2014two}
Karen Simonyan and Andrew Zisserman.
\newblock Two-stream convolutional networks for action recognition in videos.
\newblock \emph{Advances in neural information processing systems}, 27, 2014.

\bibitem[Spurlock et~al.(2024)Spurlock, Acun, Saka, and Nasraoui]{spurlock2024chatgpt}
Kyle~Dylan Spurlock, Cagla Acun, Esin Saka, and Olfa Nasraoui.
\newblock Chatgpt for conversational recommendation: Refining recommendations by reprompting with feedback.
\newblock \emph{arXiv preprint arXiv:2401.03605}, 2024.

\bibitem[Tolstikhin et~al.(2021)Tolstikhin, Houlsby, Kolesnikov, Beyer, Zhai, Unterthiner, Yung, Steiner, Keysers, Uszkoreit, et~al.]{tolstikhin2021mlp}
Ilya~O Tolstikhin, Neil Houlsby, Alexander Kolesnikov, Lucas Beyer, Xiaohua Zhai, Thomas Unterthiner, Jessica Yung, Andreas Steiner, Daniel Keysers, Jakob Uszkoreit, et~al.
\newblock Mlp-mixer: An all-mlp architecture for vision.
\newblock \emph{Advances in neural information processing systems}, 34:\penalty0 24261--24272, 2021.

\bibitem[Tran et~al.(2015)Tran, Bourdev, Fergus, Torresani, and Paluri]{tran2015learning}
Du Tran, Lubomir Bourdev, Rob Fergus, Lorenzo Torresani, and Manohar Paluri.
\newblock Learning spatiotemporal features with 3d convolutional networks.
\newblock In \emph{Proceedings of the IEEE international conference on computer vision}, pages 4489--4497, 2015.

\bibitem[Trockman and Kolter(2022)]{trockman2022patches}
Asher Trockman and J~Zico Kolter.
\newblock Patches are all you need?
\newblock \emph{arXiv preprint arXiv:2201.09792}, 2022.

\bibitem[Vadis et~al.()Vadis, Carreira, and Zisserman]{vadisaction}
Quo Vadis, Joao Carreira, and Andrew Zisserman.
\newblock Action recognition? a new model and the kinetics dataset.
\newblock \emph{Joao Carreira, Andrew Zisserman}.

\bibitem[Vaswani et~al.(2017)Vaswani, Shazeer, Parmar, Uszkoreit, Jones, Gomez, Kaiser, and Polosukhin]{vaswani2017attention}
Ashish Vaswani, Noam Shazeer, Niki Parmar, Jakob Uszkoreit, Llion Jones, Aidan~N Gomez, {\L}ukasz Kaiser, and Illia Polosukhin.
\newblock Attention is all you need.
\newblock \emph{Advances in neural information processing systems}, 30, 2017.

\bibitem[Wang et~al.(2016)Wang, Oneata, Verbeek, and Schmid]{wang2016robust}
Heng Wang, Dan Oneata, Jakob Verbeek, and Cordelia Schmid.
\newblock A robust and efficient video representation for action recognition.
\newblock \emph{International journal of computer vision}, 119:\penalty0 219--238, 2016.

\bibitem[Wang et~al.(2020)Wang, Li, Khabsa, Fang, and Ma]{wang2020linformer}
Sinong Wang, Belinda~Z Li, Madian Khabsa, Han Fang, and Hao Ma.
\newblock Linformer: Self-attention with linear complexity.
\newblock \emph{arXiv preprint arXiv:2006.04768}, 2020.

\bibitem[Wang et~al.(2021)Wang, Xie, Li, Fan, Song, Liang, Lu, Luo, and Shao]{wang2021pyramid}
Wenhai Wang, Enze Xie, Xiang Li, Deng-Ping Fan, Kaitao Song, Ding Liang, Tong Lu, Ping Luo, and Ling Shao.
\newblock Pyramid vision transformer: A versatile backbone for dense prediction without convolutions.
\newblock In \emph{Proceedings of the IEEE/CVF international conference on computer vision}, pages 568--578, 2021.

\bibitem[Wang et~al.(2022)Wang, Zhang, Guo, Dai, Chen, and Luo]{wang2022task}
Weizhi Wang, Zhirui Zhang, Junliang Guo, Yinpei Dai, Boxing Chen, and Weihua Luo.
\newblock Task-oriented dialogue system as natural language generation.
\newblock In \emph{Proceedings of the 45th International ACM SIGIR Conference on Research and Development in Information Retrieval}, pages 2698--2703, 2022.

\bibitem[Wu et~al.(2021)Wu, Xiao, Codella, Liu, Dai, Yuan, and Zhang]{wu2021cvt}
Haiping Wu, Bin Xiao, Noel Codella, Mengchen Liu, Xiyang Dai, Lu Yuan, and Lei Zhang.
\newblock Cvt: Introducing convolutions to vision transformers.
\newblock In \emph{Proceedings of the IEEE/CVF international conference on computer vision}, pages 22--31, 2021.

\bibitem[Wu et~al.(2022)Wu, Liu, Zhan, and Cheng]{wu2022p2t}
Yu-Huan Wu, Yun Liu, Xin Zhan, and Ming-Ming Cheng.
\newblock P2t: Pyramid pooling transformer for scene understanding.
\newblock \emph{IEEE Transactions on Pattern Analysis and Machine Intelligence}, 2022.

\bibitem[Yang and Tian(2014)]{yang2014super}
Xiaodong Yang and YingLi Tian.
\newblock Super normal vector for activity recognition using depth sequences.
\newblock In \emph{Proceedings of the IEEE conference on computer vision and pattern recognition}, pages 804--811, 2014.

\bibitem[Yang et~al.(2018)Yang, Molchanov, and Kautz]{yang2018making}
Xiaodong Yang, Pavlo Molchanov, and Jan Kautz.
\newblock Making convolutional networks recurrent for visual sequence learning.
\newblock In \emph{Proceedings of the IEEE Conference on Computer Vision and Pattern Recognition}, pages 6469--6478, 2018.

\bibitem[Yang et~al.(2021)Yang, Li, and Quan]{yang2021ubar}
Yunyi Yang, Yunhao Li, and Xiaojun Quan.
\newblock Ubar: Towards fully end-to-end task-oriented dialog system with gpt-2.
\newblock In \emph{Proceedings of the AAAI Conference on Artificial Intelligence}, pages 14230--14238, 2021.

\bibitem[Yu et~al.(2022)Yu, Luo, Zhou, Si, Zhou, Wang, Feng, and Yan]{yu2022metaformer}
Weihao Yu, Mi Luo, Pan Zhou, Chenyang Si, Yichen Zhou, Xinchao Wang, Jiashi Feng, and Shuicheng Yan.
\newblock Metaformer is actually what you need for vision.
\newblock In \emph{Proceedings of the IEEE/CVF conference on computer vision and pattern recognition}, pages 10819--10829, 2022.

\bibitem[Yu et~al.(2021)Yu, Zhou, Wan, Wang, Chen, Liu, Li, and Zhao]{yu2021searching}
Zitong Yu, Benjia Zhou, Jun Wan, Pichao Wang, Haoyu Chen, Xin Liu, Stan~Z Li, and Guoying Zhao.
\newblock Searching multi-rate and multi-modal temporal enhanced networks for gesture recognition.
\newblock \emph{IEEE Transactions on Image Processing}, 30:\penalty0 5626--5640, 2021.

\bibitem[Yuan et~al.(2021)Yuan, Guo, Liu, Zhou, Yu, and Wu]{yuan2021incorporating}
Kun Yuan, Shaopeng Guo, Ziwei Liu, Aojun Zhou, Fengwei Yu, and Wei Wu.
\newblock Incorporating convolution designs into visual transformers.
\newblock In \emph{Proceedings of the IEEE/CVF International Conference on Computer Vision}, pages 579--588, 2021.

\bibitem[Zamir et~al.(2022)Zamir, Arora, Khan, Hayat, Khan, and Yang]{zamir2022restormer}
Syed~Waqas Zamir, Aditya Arora, Salman Khan, Munawar Hayat, Fahad~Shahbaz Khan, and Ming-Hsuan Yang.
\newblock Restormer: Efficient transformer for high-resolution image restoration.
\newblock In \emph{Proceedings of the IEEE/CVF conference on computer vision and pattern recognition}, pages 5728--5739, 2022.

\bibitem[Zhang et~al.(2020)Zhang, Wang, and Lan]{zhang2020dynamic}
Wenjin Zhang, Jiacun Wang, and Fangping Lan.
\newblock Dynamic hand gesture recognition based on short-term sampling neural networks.
\newblock \emph{IEEE/CAA Journal of Automatica Sinica}, 8\penalty0 (1):\penalty0 110--120, 2020.

\bibitem[Zheng et~al.(2023{\natexlab{a}})Zheng, Liu, Qi, and Chen]{zheng2023potter}
Ce Zheng, Xianpeng Liu, Guo-Jun Qi, and Chen Chen.
\newblock Potter: Pooling attention transformer for efficient human mesh recovery.
\newblock In \emph{Proceedings of the IEEE/CVF Conference on Computer Vision and Pattern Recognition}, pages 1611--1620, 2023{\natexlab{a}}.

\bibitem[Zheng et~al.(2023{\natexlab{b}})Zheng, Mendieta, and Chen]{zheng2023poster}
Ce Zheng, Matias Mendieta, and Chen Chen.
\newblock Poster: A pyramid cross-fusion transformer network for facial expression recognition.
\newblock In \emph{Proceedings of the IEEE/CVF International Conference on Computer Vision}, pages 3146--3155, 2023{\natexlab{b}}.

\bibitem[Zhou et~al.(2022{\natexlab{a}})Zhou, Wang, Wan, Liang, Wang, Zhang, Lei, Li, and Jin]{zhou2022decoupling}
Benjia Zhou, Pichao Wang, Jun Wan, Yanyan Liang, Fan Wang, Du Zhang, Zhen Lei, Hao Li, and Rong Jin.
\newblock Decoupling and recoupling spatiotemporal representation for rgb-d-based motion recognition.
\newblock In \emph{Proceedings of the IEEE/CVF Conference on Computer Vision and Pattern Recognition}, pages 20154--20163, 2022{\natexlab{a}}.

\bibitem[Zhou et~al.(2021)Zhou, Kang, Jin, Yang, Lian, Jiang, Hou, and Feng]{zhou2021deepvit}
Daquan Zhou, Bingyi Kang, Xiaojie Jin, Linjie Yang, Xiaochen Lian, Zihang Jiang, Qibin Hou, and Jiashi Feng.
\newblock Deepvit: Towards deeper vision transformer.
\newblock \emph{arXiv preprint arXiv:2103.11886}, 2021.

\bibitem[Zhou et~al.(2022{\natexlab{b}})Zhou, Zhao, Wang, Wang, and Foroosh]{zhou2022centerformer}
Zixiang Zhou, Xiangchen Zhao, Yu Wang, Panqu Wang, and Hassan Foroosh.
\newblock Centerformer: Center-based transformer for 3d object detection.
\newblock In \emph{European Conference on Computer Vision}, pages 496--513. Springer, 2022{\natexlab{b}}.

\bibitem[Zhuang et~al.(2022)Zhuang, Wang, Tao, and Shang]{zhuang2022waveformer}
Yufan Zhuang, Zihan Wang, Fangbo Tao, and Jingbo Shang.
\newblock Waveformer: Linear-time attention with forward and backward wavelet transform.
\newblock 2022.

\end{thebibliography}
}


\end{document}